\documentclass{article}

\usepackage{microtype}
\usepackage{graphicx}
\usepackage{subcaption}
\usepackage[most]{tcolorbox}
\usepackage{booktabs} % for professional tables

\usepackage{hyperref}

\usepackage[accepted]{icml2026}

\usepackage{amsmath}
\allowdisplaybreaks
\usepackage{amssymb}
\usepackage{mathtools}
\usepackage{amsthm}

\usepackage{tikz}
\usetikzlibrary{arrows.meta, positioning, shapes.geometric, fit, backgrounds, calc}

\usepackage[capitalize,noabbrev]{cleveref}

\theoremstyle{plain}

\theoremstyle{definition}

\theoremstyle{remark}

\icmltitlerunning{Cumulative Prediction Error Improvement as a Tractable Intrinsic Reward}

\begin{document}

\twocolumn[
  \icmltitle{Curiosity-Critic: Cumulative Prediction Error Improvement as a \\Tractable Intrinsic Reward for World Model Training}



  \icmlsetsymbol{equal}{*}

  \begin{icmlauthorlist}
    \icmlauthor{Vin Bhaskara}{yyy}
    \icmlauthor{Haicheng Wang}{yyy}
  \end{icmlauthorlist}

  \icmlaffiliation{yyy}{Department of Computer Science, University of Toronto, Toronto, Canada}

  \icmlcorrespondingauthor{Vin Bhaskara}{bhaskara@cs.toronto.edu}
  % \icmlcorrespondingauthor{Haicheng Wang}{hcwang@cs.toronto.edu}

  \icmlkeywords{Curiosity, World Models, Aleatoric Uncertainty, Irreducible Noise, Intrinsic Reward, Reinforcement Learning, Exploration}

  \vskip 0.3in
]



\printAffiliationsAndNotice{}  % no special notice (required even if empty)

\begin{abstract}
Local prediction-error-based curiosity rewards focus on the current transition without considering the world model's cumulative prediction error across all visited transitions. We introduce \textit{Curiosity-Critic}, which grounds its intrinsic reward in the improvement of this cumulative objective, and show that it admits a tractable per-step surrogate: the difference between the current prediction error and the asymptotic error baseline of the current state transition. We estimate this error baseline online with a learned critic co-trained alongside the world model; since the critic only has to learn how hard a transition is to predict, its estimate of the irreducible noise floor converges well before the world model saturates, redirecting exploration toward learnable transitions. The reward is higher for learnable transitions and collapses toward zero for stochastic ones, thereby separating epistemic (reducible) from aleatoric (irreducible) prediction error online. Prior prediction-error curiosity formulations, from Schmidhuber (1991) to learned-feature-space variants, emerge as special cases corresponding to specific approximations of this error baseline. Experiments on a stochastic grid world show that Curiosity-Critic outperforms prediction-error, visitation-count, and Random Network Distillation methods in training speed and final world model accuracy.
\end{abstract}

\section{Introduction}

Reinforcement Learning (RL) aims to solve control problems by learning policies that maximize the expected cumulative future reward in environments with potentially complex and stochastic dynamics. Model-based RL methods learn an explicit representation of the environment -- the ``world model'' \cite{ha2018worldmodels} -- and use it to plan \cite{sutton1990dyna}. Compared to model-free approaches, model-based methods offer the promise of higher sample efficiency and a reusable world model that transfers across tasks. In practice, however, their performance is bottlenecked by the quality of the world model: an inaccurate model leads to compounding planning errors. Efficiently training a robust world model is therefore the central challenge in model-based RL.{\renewcommand{\thefootnote}{}\footnote{Code available at \url{https://github.com/vinbhaskara/Curiosity-Critic}}\renewcommand{\thefootnote}{\arabic{footnote}}}

Training a world model from trajectories sampled by a random exploration policy is highly sample inefficient. \citet{CV1,CV2} introduced curiosity-driven exploration to address this: a ``curious agent'' receives an intrinsic reward for visiting transitions that are informative for world model training. 

Schmidhuber's foundational formulations \cite{CV1,CV2} reward the agent proportionally to the model's raw prediction error \cite{CV1} and its one-step improvement \cite{CV2}, each evaluated at the most recently visited transition. A key limitation shared by both formulations is that the curiosity reward is computed solely from the \emph{current} state transition, without accounting for the world model's prediction error across the history of all previously visited transitions $\mathcal{D}_t$. This provides no incentive to visit transitions that would globally reduce the world model's cumulative error, and, therefore, can cause degradation on previously learned transitions as exploration proceeds. A second, related limitation is that neither formulation cleanly separates aleatoric (intrinsic, irreducible) from epistemic (reducible) prediction error \cite{hullermeier2021aleatoric}: \citet{CV1} makes no such distinction at all and therefore becomes trapped at aleatoric transitions such as a ``noisy TV'' \cite{pathak1_icm, pathak1}; \citet{CV2} attempts the distinction only through a noisy single-sample proxy, the post-update error, whose accuracy as an aleatoric estimate is by construction tied to the world model's own convergence.

In this work, we propose \textit{Curiosity-Critic}: a curiosity reward grounded in the world model's cumulative prediction error improvement over all visited transitions $\mathcal{D}_t$. We show that this seemingly intractable reward -- requiring world model evaluation over the entire history -- yields a tractable per-step surrogate: the difference between the model's current prediction error and its \emph{asymptotic error baseline} at convergence. To estimate the asymptotic error baseline online, we propose a learned neural critic model co-trained with the world model -- \textit{the Curiosity-Critic architecture}.

We additionally show that Schmidhuber's curiosity formulations \cite{CV1,CV2} are special cases of our framework, corresponding to specific approximations of the asymptotic error baseline. Experiments on a stochastic 2D grid world demonstrate that Curiosity-Critic consistently learns a more accurate world model with faster convergence than prediction-error, Random Network Distillation (RND) \cite{rnd}, and visitation-count methods across multiple random seeds. The grid world is a deliberately minimal controlled setting in which learnable and irreducibly unlearnable transitions coexist, so any measured advantage is attributable to the reward signal itself rather than to representation learning or architectural confounds.

\section{Previous Works}  \label{sec:background}

Intrinsic motivation as a driver of exploration has a long history in both cognitive science and machine learning \cite{oudeyer2007}. A prominent family of intrinsic curiosity rewards is based on the world model's prediction error, where the agent is rewarded for visiting state transitions that the model predicts unreliably. We review the two foundational formulations in this family, which our work directly builds upon.

\citet{CV1} proposed an intrinsic curiosity reward proportional to the model's prediction error at the current state transition. For convenience, we refer to this formulation as ``Curiosity V1'' \cite{CV1}. The reward is given by:
\begin{align}r_t(s_t,a_t) &= e(s_t,a_t~|~\theta_t) = \| \theta_{t}(s_{t}, a_{t}) - s_{t+1} \|, \label{v1eq}\end{align}
where $e(\cdot)$ denotes model error, $\|\cdot\|$ is the $L_2$ norm, and $\theta_{t}(s_{t}, a_{t})$ is the prediction of the next state $s_{t+1}$ by the world model at time $t$. This reward enables more efficient exploration than random action selection by directing the agent toward state transitions where the model is least reliable.

Throughout this paper, $\theta_t(s_t, a_t)$ denotes the world model's output at time $t$, and $\theta_t$ is used as shorthand for the world model itself. We use $e(\cdot)$ as a generic notation for the world model's prediction error on a transition, without committing to any specific form: $e(\cdot)$ may be an $L_2$ or squared $L_2$ distance on raw observations or in a learned feature space \cite{pathak1_icm}, a negative log-likelihood or KL divergence when $\theta_t$ outputs a distribution over next states, or any other consistent measure of prediction error. When we refer to ``Curiosity V1'' in the remainder of this paper, we mean the full class of intrinsic rewards $r_t = e(s_t, a_t \mid \theta_t)$ under any such choice of $e(\cdot)$; Schmidhuber's $L_2$-on-observations \cite{CV1} is the original instance, and ICM \cite{pathak1_icm} is a learned-feature-space instance of the same conceptual reward.

In general, real-world state transitions can be fundamentally non-deterministic and unlearnable. For example, consider a noisy TV \cite{pathak1} playing white noise. The next state $s_{t+1}$ of TV pixels is independent of the current state $s_t$, and hence is fundamentally unpredictable. In such environments, Curiosity V1 \cite{CV1} fails to perform better than a random policy: highly noisy state transitions constantly trigger a positive curiosity reward that does not decay with time, since the world model cannot learn the intrinsic randomness of such transitions. The curious agent becomes stuck at such high-entropy basins of attraction.

In a subsequent work \cite{CV2}, Schmidhuber proposed rewarding improvement in the world model rather than its raw error (for a broader treatment of intrinsic motivation as compression progress, see \cite{schmidhuber2010}). We refer to the one-step prediction-error instantiation of this reliability-improvement idea as ``Curiosity V2'' for brevity. In this instantiation, the decrease in prediction error after one training step forms the curiosity reward:

\begin{align} r_t(s_t, a_t) &= e(s_t, a_t ~|~\theta_t) - e(s_t, a_t~|~\theta_{t+1})\nonumber\\&=\| \theta_{t}(s_{t}, a_{t}) - s_{t+1} \|-\| \theta_{t+1}(s_{t}, a_{t}) - s_{t+1} \|,\label{v2eq}\end{align}

where $\theta_{t+1}$ is the world model after one gradient step on the transition $(s_t, a_t)\rightarrow s_{t+1}$. Both error terms are evaluated on the same transition, so the environment is sampled only once. Since noisy transitions are unlearnable, the one-step improvement $e(s_t,a_t ~|~ \theta_t) - e(s_t,a_t ~|~ \theta_{t+1})$ is smaller for such transitions in expectation, and larger for learnable ones. Unlike Curiosity V1 \cite{CV1}, which rewards raw prediction error and therefore keeps assigning high reward to irreducibly noisy transitions, Curiosity V2 \cite{CV2} naturally suppresses reward for unlearnable transitions as training proceeds.

Both Curiosity V1 \cite{CV1} and V2 \cite{CV2} compute their reward solely from the \emph{current} transition, without accounting for the model's error across the full history $\mathcal{D}_t$ of visited transitions. Consequently, neither incentivizes sampling transitions that reduce cumulative error across previously learned ones, and may fail to prevent degradation on already-learned transitions. Curiosity V1 \cite{CV1} is further compounded by using the raw prediction error, which provides no mechanism to distinguish learnable from unlearnable transitions at all. This motivates our cumulative prediction-error-based formulation in Section~\ref{sec:cumulative}.

A line of work extends Curiosity V1 \cite{CV1} by varying the choice of $e(\cdot)$: ICM \cite{pathak1_icm} measures squared $L_2$ error in a learned inverse-dynamics feature space, and \citet{pathak1} provide a large-scale empirical study of feature-space choices. These inherit Curiosity V1's per-step structure and can still get stuck at unlearnable, stochastic transitions. Other approaches avoid raw observation error: Random Network Distillation \cite{rnd} uses a fixed random target to produce a soft count-based bonus that is less tied to transition prediction error; Disagreement \cite{disagreement} uses ensemble variance to be noise-robust at the cost of training and querying multiple world models, making the approach substantially more expensive computationally than single-model curiosity objectives. VIME \cite{vime} and count-based bonuses \cite{bellemare2016} take Bayesian and visitation perspectives. 

Our study is deliberately scoped to intrinsic rewards computable from a single world model: every method trains the same world-model architecture and differs only in the reward signal, so any performance gap traces to the reward formulation itself, not to representation or architecture. Two families fall outside this single-model scope. Ensemble or Bayesian methods such as Disagreement \cite{disagreement} need multiple world models, a different setting rather than a single-model reward. Feature-space variants such as ICM change the representation, an axis we hold fixed across all methods. Both compose with the Curiosity-Critic rather than competing with it. We therefore include Curiosity V1 and exact visitation counts as baselines, since ICM is a feature-space instance of Curiosity V1 and exact counts are the tabular form of pseudo-count bonuses.

\section{Cumulative Prediction Error Improvement as an Intrinsic Reward}\label{sec:cumulative}

We propose a curiosity reward based on the world model's total prediction error improvement across all learnable state transitions visited up to time $t$. Consider the following curiosity reward $r_t$ at time $t$ that sums up the changes in prediction errors after a training update over all state transitions $\mathcal{D}_t=\{ s_0,a_0,s_1,a_1,\dots,s_{t},a_t\}$ up to time $t$: 
\begin{align}
&r_t(s_t, a_t~|~\mathcal{D}_t)\nonumber \\&= \sum \limits_{\substack{t' \leq t\\(s_{t'},a_{t'}) \in \mathcal{D}_t}} \left\{ e(s_{t'}, a_{t'}~|~\theta_t) - e(s_{t'}, a_{t'}~|~\theta_{t+1}) \right\}, \label{formulation1}
\end{align}
where $e(s_{t'}, a_{t'}~|~\theta_t) = \| \theta_{t}(s_{t'}, a_{t'}) - s_{t'+1} \|$ is the model error in predicting a state transition sampled previously at time $t'$ from the history $\mathcal{D}_t$ with the model $\theta_t$ at time $t$. Note that in this notation, $s_{t'+1}$ corresponds to the true state observed at time $t'+1$, and $\theta_t(s_{t'},a_{t'})$ corresponds to the prediction for the next state at $t'+1$ by the world model at time $t$. Also note that $\|\cdot\|$ represents the $L_2$ norm above as the state $s_t$ can generally be a multi-dimensional vector.

Splitting $\mathcal{D}_t=\mathcal{D}_t^+ \cup \mathcal{D}_t^-$ into learnable and unlearnable subsets, the contribution from $\mathcal{D}_t^-$ to Eq.~\eqref{formulation1} vanishes in expectation (a one-step update on the current transition produces zero expected error change on purely noisy transitions). Thus Eq.~\eqref{formulation1} measures, in expectation, the cumulative error improvement across all \emph{learnable} transitions in $\mathcal{D}_t$ after the update $\theta_t \rightarrow \theta_{t+1}$ (see Appendix~\ref{app:learnablesplit}).

\subsection{Tractable Per-Step Reformulation}\label{sec:subsectractableref}

Evaluating Eq.~\eqref{formulation1} at each step is prohibitive: it requires running the world model over all of $\mathcal{D}_t$, costing $\mathcal{O}(t^2)$ compute and $\mathcal{O}(t)$ memory. We now derive an approximate tractable per-step form.

Consider the cumulative future reward $C(T) = \sum_{t=0}^{T}\gamma^t ~ r_t(s_t, a_t~|~\mathcal{D}_t)$ that the agent maximizes over $T$ steps:
\begin{align}
C(T) &= \sum \limits_{t=0}^{T} \gamma^t~\sum \limits_{t' \leq t} \left\{ e(s_{t'}, a_{t'}~|~\theta_t) - e(s_{t'}, a_{t'}~|~\theta_{t+1}) \right\}, \label{telescoping}
\end{align}
where $(s_{t'},a_{t'})\in\mathcal{D}_t$. This cumulative sum decomposes exactly into an intractable history term and a tractable per-step term (full derivation in Appendix~\ref{app:derivation}):
\begin{align}
&C(T) =\underbrace{\left(1-\tfrac{1}{\gamma}\right) ~  \sum \limits_{t=0}^{T} ~~ \big\{\gamma^{t}\sum \limits_{t' < t} e(s_{t'}, a_{t'}~|~\theta_{t})\big\}}_{\text{intractable history term}} ~~~~~~~~+\nonumber\\  &\underbrace{\left(\tfrac{1}{\gamma}\right) ~ \sum \limits_{t=0}^{T} ~~ \big\{\gamma^{t+1}  ~e(s_t,a_t~|~\theta_t) - \gamma^{T+1}~e(s_t,a_t~|~\theta_{T+1}) \big\}}_{\text{tractable per-step term}}. \label{eqCT18}
\end{align}
The agent's goal is end-of-training model accuracy, not near-term reward, which motivates the limit $\gamma\to 1$. Under this limit, the intractable history term vanishes (it is negative for $\gamma\in(0,1)$, so the tractable per-step term is an upper bound on $C(T)$ that becomes tight as $\gamma\to 1$; see Appendix~\ref{app:upperbound}). Setting $\gamma\to 1$ in Eq.~\eqref{eqCT18} collapses the expression to a per-step telescoping sum in the model parameters (see Appendix~\ref{app:telescoping}):
\begin{align}
C(T) \big|_{\gamma \rightarrow 1} &= \sum \limits_{t=0}^{T} \bigg[e(s_t,a_t~|~\theta_t) ~- ~e(s_t,a_t~|~\theta_{T+1}) \bigg]\label{wow1}.
\end{align}
This identifies an effective per-step reward,
\begin{align}
r^{\text{eff}}_t(s_t,a_t) = e(s_t, a_t ~|~\theta_t) - e(s_t, a_t~|~\theta_{T+1}), \label{effrewardv3}
\end{align}
whose sum, $\sum_{t=0}^{T}\gamma^t ~r^{\text{eff}}_t(s_t,a_t)|_{\gamma\to 1}$, exactly reproduces $C(T)|_{\gamma\to 1}$ without any approximations. The identified effective per-step reward $r^{\text{eff}}_t(s_t,a_t)$ is a standalone signal that any RL algorithm can consume with its own discount factor; the $\gamma\to 1$ limit is only used to derive the per-step form, not to constrain the downstream optimizer.

The remaining challenge is estimating $e(s_t,a_t~|~\theta_{T+1})$ online. Two approximations bridge this gap: \textbf{(a) Horizon Approximation.} Replace $\theta_{T+1}\to\theta_\infty$, the asymptotic convergence limit (exact as $T\to\infty$); \textbf{(b) Sample-to-Expectation Approximation.} Replace the single-sample error with its expectation over the state transition probability $\mathcal{P}(\cdot | s_t, a_t)$, given by $e(s_t,a_t~|~\theta_\infty) \to \mathbb{E}_\mathcal{P}[e(s_t,a_t~|~\theta_\infty)]$. This is a variance-reduction step that makes the error baseline amenable to online learning (Section~\ref{sec:background_critic}). Together they yield the final form of our proposed per-step tractable reward:
\begin{equation}
    r^{\text{eff}}_t(s_t,a_t) \approx e(s_t, a_t ~|~\theta_t) - \mathbb{E}_{\mathcal{P}}\left[ e(s_t, a_t~|~\theta_{\infty})\right]. \label{effrewardv3opt}
\end{equation}

\begin{tcolorbox}[colback=gray!5, colframe=gray!55, boxrule=0.4pt, arc=2pt, left=5pt, right=5pt, top=4pt, bottom=4pt]
\textbf{Key insight.} In the $\gamma\to 1$ limit, the cumulative error improvement over the entire history $\mathcal{D}_t$ after a training step, $r_t(s_t, a_t~|~\mathcal{D}_t)$, is approximated in expectation by a single per-step quantity $r^{\text{eff}}_t(s_t,a_t)$: the error on the \emph{current} transition \mbox{$(s_t,a_t)\to s_{t+1}$} in excess of its asymptotic baseline.
\end{tcolorbox}

\textbf{Schmidhuber's formulations as special cases.} Our framework reveals Curiosity V1 \cite{CV1} and V2 \cite{CV2} as two points on a single axis: the same proposed per-step tractable reward (Eq.~\eqref{effrewardv3opt}), approximated by setting the asymptotic error baseline to zero (Curiosity V1 \cite{CV1}, assuming a fully deterministic environment) or to a one-step post-update proxy $e(s_t, a_t~|~\theta_{t+1})$ (Curiosity V2 \cite{CV2}). Since these reductions hold for any valid $e(\cdot)$, learned-feature-space variants of Curiosity V1 such as ICM \cite{pathak1_icm} are likewise special cases of our framework under the zero-error-baseline approximation. Curiosity-Critic replaces this approximation with a learned estimator (Section \ref{sec:estimationoferrbaseline}). Further details and intuition  are discussed in Appendix~\ref{app:specialcases}.

\section{Estimating the Asymptotic Error Baseline}\label{sec:estimationoferrbaseline}

In this section, we show how $\mathbb{E}_{\mathcal{P}}\left[ e(s_t, a_t~|~\theta_{\infty})\right]$ can be estimated online during world model training, and introduce the Curiosity-Critic framework that implements this estimation.

\subsection{Asymptotic Error Baseline: The Irreducible Noise Floor}

Trained to minimize a mean squared error (MSE) loss, a converged world model's prediction approaches the conditional mean, $\theta_\infty(s_t,a_t)\to \mathbb{E}_\mathcal{P}[s_{t+1}|s_t,a_t]$ (other losses induce different statistics). The asymptotic error baseline therefore equals the mean absolute deviation of $s_{t+1}$ under $\mathcal{P}(\cdot|s_t,a_t)$,
\begin{align}
\mathbb{E}_{\mathcal{P}}\left[ e(s_t, a_t~|~\theta_{\infty})\right]= \mathbb{E}_{\mathcal{P}}\left[ \| \mathbb{E}_{\mathcal{P}}\left[ s_{t+1}|s_t,a_t \right] - s_{t+1} \| \right], \label{opterrorform}
\end{align}
which is precisely the irreducible noise floor that no model can learn away. Subtracting it from $e(s_t,a_t~|~\theta_t)$ leaves only the \emph{learnable} component as the reward $r^{\text{eff}}_t$. In the fully stochastic limit the current error $e(s_t,a_t~|~\theta_t)$ approaches the error baseline as the training proceeds and $r^{\text{eff}}_t\to 0$; in the fully deterministic limit the error baseline is zero and $r^{\text{eff}}_t = e(s_t,a_t~|~\theta_t)$; for intermediate transitions $r^{\text{eff}}_t$ tracks the epistemic error. A Jensen-inequality variance upper bound is given in Appendix~\ref{app:jensen}.

Under the idealization that $\theta_\infty$ attains the Bayes-optimal conditional mean, the asymptotic error baseline corresponds exactly to the aleatoric component of prediction error, and subtracting it leaves the reducible, epistemic component that should guide exploration. With finite capacity and finite-time optimization, $\theta_\infty$ is instead the best model this architecture and optimizer actually reach, so the baseline also absorbs the epistemic error this model cannot reduce. Subtracting it then isolates the error the current model can still drive down, steering exploration toward transitions that are learnable not just in principle but in practice -- an efficient use of finite capacity.

\subsection{Estimating the Baseline Online: the Curiosity-Critic}\label{sec:background_critic}

The quantity $\mathbb{E}_{\mathcal{P}}\left[ e(s_t, a_t~|~\theta_{\infty})\right]$ is not directly observable, since neither $\mathcal{P}$ nor $\theta_\infty$ are known.

At each time step $t$, however, the agent observes a single sample $s_{t+1}$ from the environment. This observation is used to update the world model from $\theta_t$ to $\theta_{t+1}$, after which the post-update error $e(s_t, a_t~|~\theta_{t+1})$ -- the error of the updated model on the same transition -- can be computed at no additional cost. This serves as a natural single-sample estimate of the asymptotic baseline available at time $t$: as $\theta_t \rightarrow \theta_\infty$, the post-update error converges to the irreducible noise floor of the transition.

One could use this single-sample estimate directly as the baseline at each step -- corresponding to the one-step Curiosity V2 \cite{CV2} reward -- but this has two drawbacks. First, it is noisy -- a single sample is a high-variance estimate of $\mathbb{E}_{\mathcal{P}}\left[ e(s_t, a_t~|~\theta_{\infty})\right]$. Second, since the post-update error only approaches the asymptotic baseline as $\theta_{t+1} \to \theta_\infty$, the accuracy of this baseline is by construction tied to the world model's own convergence.

A better approach is to learn to predict $\mathbb{E}_{\mathcal{P}}\left[ e(s_t, a_t~|~\theta_{\infty})\right]$ directly from $(s_t, a_t)$, using $e(s_t, a_t~|~\theta_{t+1})$ as the regression target at each step. Such a learned estimator generalises across similar transitions (pooling information across states). Arguably, regressing the world model's scalar error is a simpler learning problem than predicting the next state itself: the critic only has to learn how hard a transition is to predict, not what comes next. We therefore expect the critic to converge well before the world model itself, which we verify empirically in Section~\ref{sec:experiments}. This early convergence lets the critic guide exploration toward learnable transitions ahead of world-model convergence, improving both convergence speed and final accuracy.

\subsection{The Curiosity-Critic Framework}

\begin{figure}[!t]
\centering
\resizebox{0.46\textwidth}{!}{%
\begin{tikzpicture}[
  every node/.style={font=\small},
  mbox/.style={draw, thick, minimum width=3.6cm, minimum height=1.8cm,
               align=center, rounded corners=3pt},
  abox/.style={draw, thick, minimum width=3.0cm, minimum height=0.85cm,
               align=center, rounded corners=3pt},
  wout/.style={draw=blue!60!black, thick, dashed,
               minimum width=3.0cm, minimum height=0.72cm,
               align=center, rounded corners=2pt, inner sep=4pt, fill=white},
  nout/.style={draw=orange!70!black, thick, dash dot,
               minimum width=3.0cm, minimum height=0.72cm,
               align=center, rounded corners=2pt, inner sep=4pt, fill=white},
  arr/.style={-{Stealth[scale=1.0]}, thick},
  darr/.style={-{Stealth[scale=1.0]}, thick, dashed, color=gray!50},
  num/.style={font=\scriptsize\bfseries, circle, draw=black!70, fill=white,
              inner sep=0pt, minimum size=14pt},
  lbl/.style={font=\footnotesize, fill=white, inner sep=1.5pt},
]

\node[mbox] (wm) at (-3, 0) {\textbf{World Model}\\[3pt]$\theta$};
\node[mbox] (nc) at ( 3, 0) {\textbf{Neural Critic}\\[3pt]$\phi$};

\draw[arr] (-3, -1.8) -- (wm.south)
  node[lbl, right, pos=0.5] {$s_t,\,a_t$};
\draw[arr] ( 3, -1.8) -- (nc.south)
  node[lbl, right, pos=0.5] {$s_t,\,a_t$};

\node[wout] (wm_out) at (-3,  3.15) {$e(s_t,a_t\!\mid\!\theta_t)$};
\node[nout] (nc_out) at ( 3,  3.15) {$\phi_{t+1}(s_t,a_t)$};
\node[font=\footnotesize, text=blue!60!black] at (-3, 3.68) {world model error};
\node[font=\footnotesize, text=orange!70!black] at ( 3, 3.68) {curiosity-critic};

\draw[arr] (-3.5, 0.9) -- (-3.5, 1.7);
\node[font=\footnotesize, fill=white, inner sep=1.5pt] (wmpred) at (-3.5, 1.95) {$\theta_t(s_t,a_t)$};
\draw[arr] (-3.5, 2.2) -- (-3.5, 2.8);
\node[num] at (-3.95, 1.2) {\scriptsize 1};

\draw[darr] (-2.45, 2.15) -- (-2.45, 0.9);
\node[lbl, right] at (-2.45, 1.5) {$s_{t+1}$};
\node[num] at (-1.85, 2.2) {\scriptsize 2};

\draw[darr] (3.55, 2.15) -- (3.55, 0.9);
\node[lbl, right] at (3.55, 1.5) {$e(s_t,a_t\!\mid\!\theta_{t+1})$};
\node[num] at (4.15, 2.2) {\scriptsize 3};

\draw[arr] (2.5, 0.9) -- (2.5, 2.8);
\node[num] at (2.1, 1.55) {\scriptsize 4};

\node[font=\small] at (-4.5, -3.2) {$r_t\;=$};
\node[wout] (wm_eq) at (-1.8, -3.2) {$e(s_t,a_t\!\mid\!\theta_t)$};
\node[font=\small]  at ( 0.3, -3.2) {$-$};
\node[nout] (nc_eq) at ( 2.5, -3.2) {$\phi_{t+1}(s_t,a_t)$};
\node[num] at (-4.5, -3.9) {\scriptsize 5};

\node[abox] (ag) at (0, -5.8)
  {Curiosity Agent\\[2pt]\small(updates $\pi_t$ and samples next action)};
\draw[darr] (0, -4.6) -- (ag.north)
  node[lbl, right, pos=0.5, text=black] {$r_t$};
\node[num] at (0.6, -4.7) {\scriptsize 6};

\end{tikzpicture}%
}
\caption{\textit{The Curiosity-Critic architecture.} Solid black arrows: forward-pass flow. Dashed gray arrows: backward-pass training signals. Steps: (1) world model computes error $e(s_t,a_t\!\mid\!\theta_t)$; (2) world model updates to $\theta_{t+1}$ on $s_{t+1}$; (3) critic regresses onto the post-update error $e(s_t,a_t\!\mid\!\theta_{t+1})$; (4) critic outputs baseline $\phi_{t+1}(s_t,a_t)$; (5) reward $r_t = e(s_t,a_t\!\mid\!\theta_t) - \phi_{t+1}(s_t,a_t)$; (6) the curiosity agent updates $\pi_t$ and samples the next action.}
\label{arch}
\end{figure}
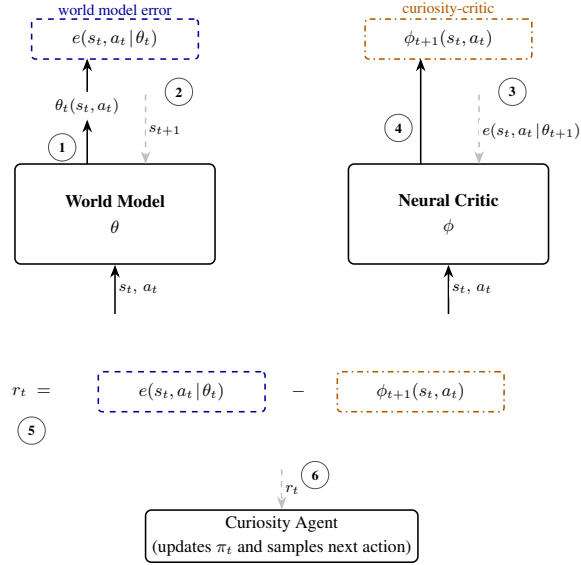

We implement this idea as the \textit{Curiosity-Critic} framework, summarised in Algorithm~\ref{alg:cc} (Appendix~\ref{app:algorithm}) and illustrated in Figure~\ref{arch}. A Neural Critic network $\phi$ is trained in parallel with the world model $\theta$. At each step $t$: \textbf{(1)} the world model $\theta_t$ computes the raw prediction $\theta_t(s_t, a_t)$ and the resulting prediction error $e(s_t, a_t~|~\theta_t)$; \textbf{(2)} the world model is updated to $\theta_{t+1}$ using the observed $s_{t+1}$; \textbf{(3)} the post-update error $e(s_t, a_t~|~\theta_{t+1})$ is used as the training target for the critic model $\phi_t$; \textbf{(4)} the updated critic $\phi_{t+1}$ produces the error baseline estimate $\phi_{t+1}(s_t, a_t)$; \textbf{(5)} the curiosity reward is formed as:
\begin{equation}
r_t(s_t, a_t) = e(s_t, a_t~|~\theta_t) ~-~ \phi_{t+1}(s_t, a_t).
\end{equation}
This reward is positive when the current model error exceeds the estimated noise floor -- i.e., when the transition is still learnable -- and near zero for stochastic, unlearnable transitions. The reward is then used as an RL signal \textbf{(6)} to update the curiosity agent's policy $\pi_t$, which selects the next action.

The framework is agnostic to the error metric: the critic regresses the asymptotic value of whatever $e(\cdot)$ is in use, so the method ports directly to feature-space formulations (ICM \cite{pathak1_icm}) or distributional world models (where the critic regresses the conditional entropy of $\mathcal{P}(\cdot \mid s_t, a_t)$). Only $e(\cdot)$ changes; Eq.~\eqref{effrewardv3opt} and Algorithm~\ref{alg:cc} (Appendix~\ref{app:algorithm}) remain unchanged. It is also agnostic to batch size and update cadence: Algorithm~\ref{alg:cc} (Appendix~\ref{app:algorithm}) is stated per environment step for clarity, but steps~(1)--(5) hold unchanged when $\theta$ and $\phi$ are updated on minibatches, with the reward optionally computed from frozen snapshots of $\theta,\phi$ before the update, as in on-policy deep RL.

Notably, training the critic and policy concurrently creates a self-correcting feedback loop: if the critic underestimates the noise floor for a stochastic transition, $r_t$ remains artificially high, causing the policy to keep revisiting that transition. Each revisit provides an additional training sample for the critic, driving $\phi_{t+1}(s_t, a_t)$ upward until it accurately reflects the irreducible error. Once it does, $r_t$ drops to near zero and the policy is redirected toward genuinely learnable transitions.

\section{Experiments}\label{sec:experiments}

\begin{table*}[t]
    \centering
    \begin{tabular}{lcccccc}
        \toprule
        Method & Seed 1 & Seed 2 & Seed 3 & Seed 4 & Seed 5 & Mean $\pm$ Std \\
        \midrule
        Random                              & $2.268$ & $2.103$ & $2.027$ & $2.263$ & $3.080$ & $2.348 \pm 0.377$ \\
        Curiosity V1 \cite{CV1}             & $7.287$ & $7.054$ & $7.093$ & $7.256$ & $6.882$ & $7.114 \pm 0.147$ \\
        Curiosity V2 \cite{CV2}             & $2.546$ & $3.128$ & $3.204$ & $3.422$ & $2.393$ & $2.939 \pm 0.398$ \\
        Visitation Count \cite{Visitation}  & $5.676$ & $5.693$ & $4.578$ & $5.052$ & $6.939$ & $5.588 \pm 0.794$ \\
        RND (State) \cite{rnd}              & $2.427$ & $2.188$ & $2.217$ & $2.131$ & $2.135$ & $2.220 \pm 0.109$ \\
        RND (Observation) \cite{rnd}              & $6.898$ & $6.810$ & $6.451$ & $7.018$ & $7.034$ & $6.842 \pm 0.212$ \\
        Ours (Tabular Critic)               & \underline{$1.921$} & \underline{$2.025$} &  \underline{$1.827$} & \underline{$1.938$} & \underline{$1.851$} & \underline{$1.912 \pm 0.070$} \\
        Ours (Neural Critic Model)         & \textbf{1.755} & \textbf{1.817} & \textbf{1.926} & \textbf{1.975} & \textbf{1.820} & \textbf{1.858} $\boldsymbol{\pm}$ \textbf{0.080} \\
        \textit{Ours Oracle (Ground-Truth Critic)} & $\mathit{1.624}$ & $\mathit{1.733}$ & $\mathit{1.816}$ & $\mathit{1.753}$ & $\mathit{1.757}$ & $\mathit{1.736} \pm \mathit{0.063}$ \\
        \bottomrule
    \end{tabular}
    \vspace{0.5em}
    \caption{Final mean L2 prediction error on deterministic cells after 35{,}000 environment steps, averaged over five independent seeds (lower is better). \textbf{Bold} denotes the best non-oracle result and \underline{underline} denotes the second-best. \textit{Italicised row} uses privileged knowledge of the environment's noise floor and is included as oracle reference.}
    \label{tab:final_error}
\end{table*}

\textbf{Environment.} The experiment is a controlled $30 \times 30$ grid world that instantiates the canonical ``noisy TV'' problem \cite{pathak1} -- a minimal setting in which learnable and irreducibly unlearnable transitions coexist by construction. The grid is split into two equal halves. The \emph{deterministic half} (columns 0--14, $450$ cells) emits, at each cell, a fixed $200$-dimensional binary ``TV-pixel'' observation, identical across visits and learnable in principle; neighbouring cells are cyclic shifts of one another, so the world model can generalize across them. The \emph{stochastic half} (columns 15--29) re-samples an i.i.d.\ Bernoulli$(0.5)$ $200$-dimensional binary observation on every visit, so prediction error there is irreducible regardless of training (analytical noise floor $\sqrt{200}\times 0.5 \approx 7.07$). The agent starts at the boundary cell $(15,15)$ on the stochastic side and selects one of four directional actions per step (Up/Down/Left/Right); cell-to-cell transitions are themselves deterministic, so all of the environment's stochasticity lives in the observations emitted by stochastic-half cells. The world model takes a $60$-dimensional factored one-hot row/column encoding of the current cell and predicts that cell's $200$-dimensional TV-pixel observation. Our primary metric is the world model's mean L2 prediction error on the $450$ deterministic cells, queried directly every $100$ environment steps (lower is better). See Appendix~\ref{app:envdetails} for further details.

\textbf{Baseline methods.} We evaluate Curiosity-Critic against Curiosity V1 \cite{CV1}, Curiosity V2 \cite{CV2}, RND \cite{rnd}, Visitation Count \cite{Visitation}, and undirected Random exploration. Because our cell coordinates are deterministic but the TV-pixel observations they emit are stochastic on the right half, we report two RND variants -- RND (State), which receives the $60$-dimensional factored row/column state, and RND (Observation), which receives the $200$-dimensional TV-pixel observation -- to separate state-novelty from observation-novelty as the input signal to RND.
Full training details (model architecture, policy, reward clipping and normalization, RND variants, oracle reference) are given in Appendix~\ref{app:training}. All experiments run for $35{,}000$ environment steps across five independent seeds, and all results are fully reproducible from the released code at \url{https://github.com/vinbhaskara/Curiosity-Critic}.

\subsection{Results}

\begin{figure*}[!t]
    \centering
    \includegraphics[width=0.796\textwidth]{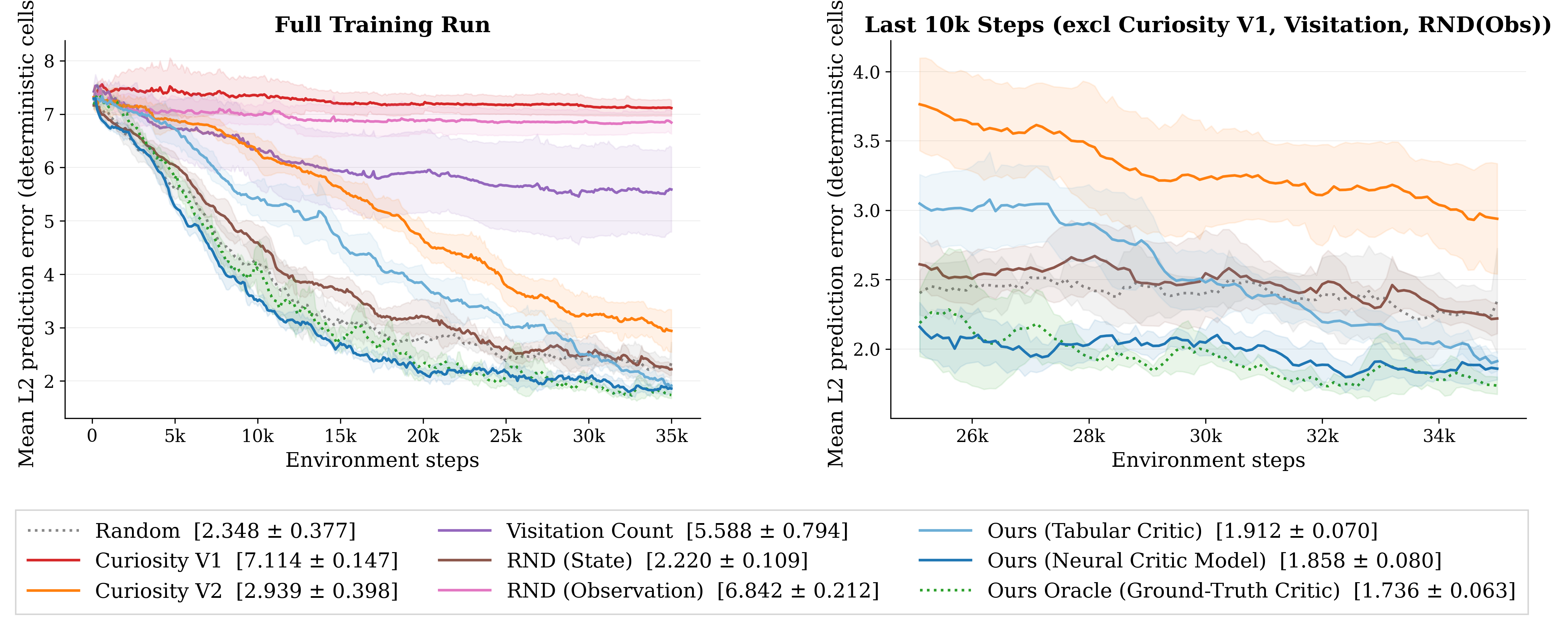}
    \caption{Mean L2 prediction error on deterministic cells versus environment steps, averaged over five seeds with $\pm 1$ standard deviation shading.
    Curiosity V1 \cite{CV1} uses raw prediction error as the intrinsic reward; Curiosity V2 \cite{CV2} uses the one-step reduction in prediction error.
    RND (State) \cite{rnd} uses the factored grid state as input, while RND (Observation) uses the stochastic TV-pixel observation.
    \textit{Left}: full 35{,}000-step training run across all nine methods.
    \textit{Right}: final 10{,}000 steps (steps 25k--35k), excluding Curiosity V1 \cite{CV1}, Visitation Count \cite{Visitation}, and RND (Observation).}
    \label{fig:error}
\end{figure*}

\begin{figure*}[!t]
    \centering
    \includegraphics[width=0.796\textwidth]{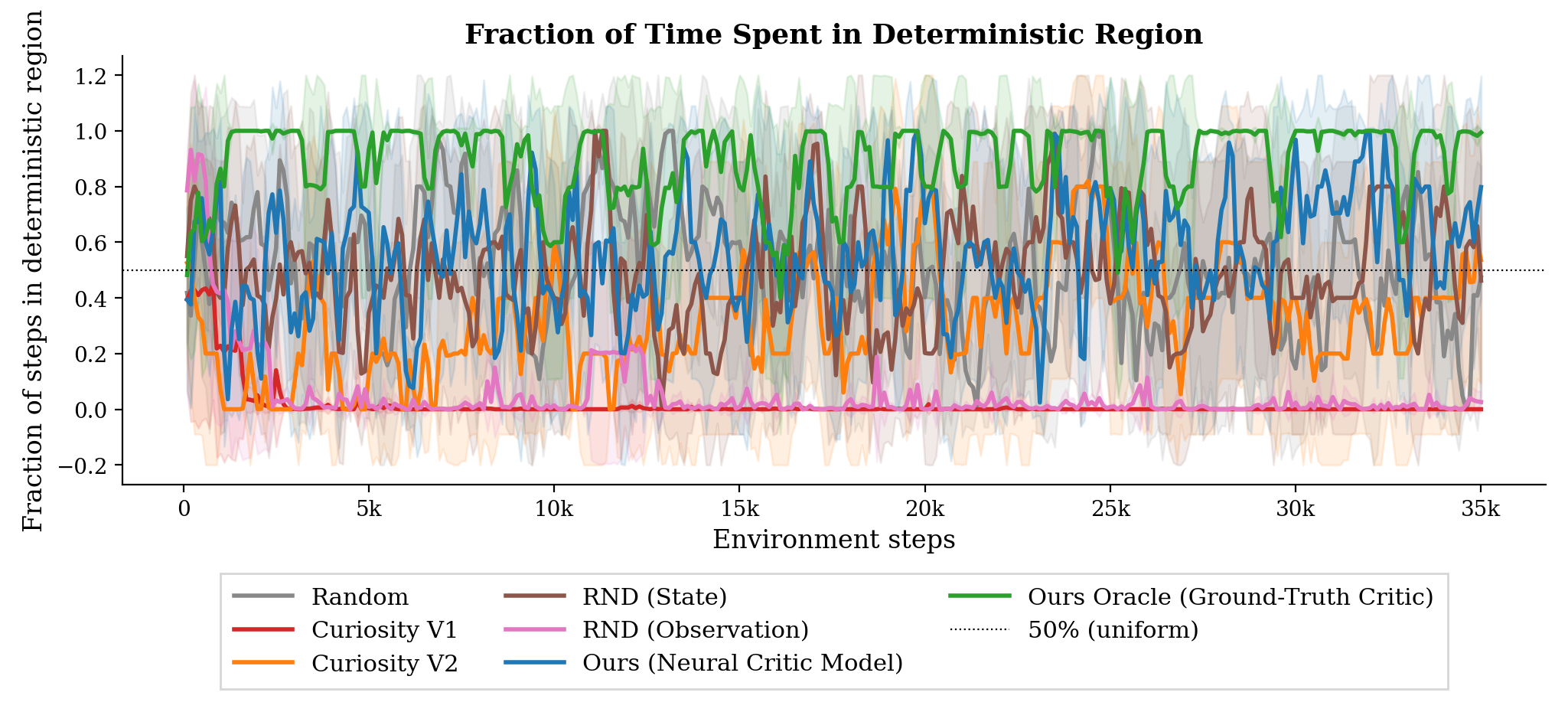}
    \caption{Fraction of environment steps spent in the deterministic region (columns 0--14) within each non-overlapping 100-step window, averaged over five seeds with $\pm 1$ standard deviation shading.
    Curiosity V1 \cite{CV1} (prediction error as reward) collapses to zero within the first few thousand steps as it becomes entirely absorbed by stochastic cells.
    Ours (Neural Critic Model) builds up a consistently high deterministic-region fraction throughout training ($70.9\%$ in the final 5{,}000 steps on average), closely tracking the oracle critic's coverage ($95.3\%$), without any privileged knowledge of cell type.
    RND (Observation) collapses toward the stochastic region like Curiosity V1, while RND (State) behaves more like a soft state-count bonus and maintains broader coverage without matching the critic's learnability-directed focus.}
    \label{fig:visitfrac}
\end{figure*}

Figure~\ref{fig:error} shows mean L2 error on deterministic cells over training (right panel zooms into steps 25k--35k, excluding Curiosity V1, RND (Observation) and Visitation Count to resolve converging methods); visit fractions are shown in Figure~\ref{fig:visitfrac}; final errors are in Table~\ref{tab:final_error}. Complementary analysis and plots, including trajectory snapshots (Figure~\ref{fig:snapshot}), and visitation heatmaps (Figures~\ref{fig:heatmap_end}, \ref{fig:heatmap_windows}), are in Appendix~\ref{app:figures}; a real-time animation is at \href{https://youtu.be/hHSvQGaO5yY}{\texttt{youtu.be/hHSvQGaO5yY}}.

\textbf{Main findings.}
Curiosity-Critic gives the best non-oracle final error: the Neural Critic reaches $\mathbf{1.858 \pm 0.080}$, and the Tabular Critic ablation reaches $\underline{1.912 \pm 0.070}$, beating every non-critic baseline including RND (State) ($2.220 \pm 0.109$). The Neural Critic is also fastest, dropping below error $3.0$ at step $\mathbf{13{,}400}$, compared with step $21{,}700$ for RND (State), $25{,}300$ for the Tabular Critic, and $34{,}400$ for Curiosity V2. It spends $70.9\%$ of the final $5{,}000$ steps in the deterministic half of the grid, compared with $52.8\%$ for RND (State) and $42.6\%$ for Random. The Neural Critic is able to generalize across states due to the input one-hot cell encoding, outperforming the Tabular Critic.

The baseline methods reveal the failure modes the Curiosity-Critic framework is designed to avoid. Curiosity V1 and RND (Observation) are drawn to the stochastic cells, ending with high final errors ($7.114 \pm 0.147$ and $6.842 \pm 0.212$), suffering complete noisy-TV failure. Curiosity V2 substantially improves over Curiosity V1 but remains a high-variance single-sample proxy for the asymptotic error baseline (Section~\ref{sec:cumulative}), finishing worse than Random on average ($2.939 \pm 0.398$ vs.\ $2.348 \pm 0.377$). Visitation Count, despite spreading coverage broadly regardless of cell learnability, plateaus at $5.588 \pm 0.794$ -- the highest seed variance among all methods. That Curiosity V2, Visitation Count, and RND (Observation) all finish at or worse than undirected random exploration -- a strong baseline here, since the grid has no interior obstacles and an undirected random walk diffuses into the learnable region with high probability -- shows that naive curiosity signals can actively harm learning under irreducible stochasticity. 

The oracle critic, which uses privileged knowledge of the true noise floor, reaches $1.736 \pm 0.063$ with $95.3\%$ late-training deterministic focus. The learned Neural Critic reaches within $5\%$ of the oracle noise floor on stochastic cells ($\sqrt{200}\times 0.5 \approx 7.07$) by step $400$ and oscillates closely around the oracle thereafter. For deterministic cells, the critic estimate rises in the first $\sim\!500$ steps (tracking the still-large post-update error early in world-model training), and then declines alongside the world model's improving accuracy. This supports our claim in Section~\ref{sec:background_critic}: the critic estimate for stochastic cells first reaches within $5\%$ of the oracle baseline by $\sim\!1\%$ of total training -- well before the world model itself converges -- confirming that the critic converges meaningfully earlier than the world model. Figure~\ref{fig:critic_convergence} (Appendix~\ref{app:figures}) shows the mean neural critic estimate over all deterministic and all stochastic cells throughout training. Detailed per-method analyses are in Appendix~\ref{app:permethod}.

\section{Conclusions}

We introduced Curiosity-Critic, a curiosity framework grounded in the world model's cumulative prediction error improvement. Our central claim is that this cumulative objective -- crediting each visited transition with the error reduction its update produces across the entire history, not just the current step -- is the right signal for training a world model, and that it collapses into a tractable per-step surrogate: the current error minus a learned asymptotic error baseline. Optimizing it targets exactly what model-based RL needs -- a more accurate world model, in fewer steps -- which our experiments on a stochastic 2D grid world confirm: Curiosity-Critic converges faster and to lower final error than prediction-error, RND, and count-based baselines, approaching oracle performance with no privileged knowledge of the noise floor. Cleanly separating reducible (epistemic) from irreducible (aleatoric) error, and so resolving the noisy-TV failure, follows as a consequence of this objective.

\textbf{Limitations and Future Work.} Current experiments are on a controlled grid world; behaviour in high-dimensional settings is not yet characterized empirically. A next step is evaluating Curiosity-Critic in large-scale environments such as MuJoCo \cite{todorov2012mujoco}, Atari \cite{ale}, and VizDoom \cite{vizdoom}. Finally, \emph{how} best to estimate the asymptotic error baseline remains open: richer critic inputs (e.g., the current world-model error) or a world-model--critic shared backbone may yield more accurate baseline estimates.

\section*{Acknowledgments}
We are thankful to Prof. Jimmy Ba at the University of Toronto for his comments and feedback during the early stages of this work.

\section*{Impact Statement}

This paper presents work whose goal is to advance Machine
Learning. By improving an agent's ability to distinguish reducible from irreducible model error, this work can support efficient exploration and training of world models.


\bibliography{example_paper}
\bibliographystyle{icml2026}

%%%%%%%%%%%%%%%%%%%%%%%%%%%%%%%%%%%%%%%%%%%%%%%%%%%%%%%%%%%%%%%%%%%%%%%%%%%%%%%
%%%%%%%%%%%%%%%%%%%%%%%%%%%%%%%%%%%%%%%%%%%%%%%%%%%%%%%%%%%%%%%%%%%%%%%%%%%%%%%
% APPENDIX
%%%%%%%%%%%%%%%%%%%%%%%%%%%%%%%%%%%%%%%%%%%%%%%%%%%%%%%%%%%%%%%%%%%%%%%%%%%%%%%
%%%%%%%%%%%%%%%%%%%%%%%%%%%%%%%%%%%%%%%%%%%%%%%%%%%%%%%%%%%%%%%%%%%%%%%%%%%%%%%
\newpage
\appendix
\onecolumn

\section{Learnable vs Unlearnable Decomposition}\label{app:learnablesplit}

Let $\mathcal{D}_t^+$ and $\mathcal{D}_t^-$ denote the subsets of learnable and highly noisy state transitions in the transition history $\mathcal{D}_t$ up to time $t$, respectively, so $\mathcal{D}_t=\mathcal{D}_t^+ \cup \mathcal{D}_t^-$. Let $\mathbb{E}[\cdot]$ denote the expectation over all trajectories. Then, for our proposed cumulative error improvement reward in Eq. \eqref{formulation1}, $\mathbb{E}~[r_t(s_t, a_t~|~\mathcal{D}_t)] \approx \mathbb{E}~[r_t(s_t, a_t~|~\mathcal{D}_t^+)]$, or
\begin{align}
 \mathbb{E}~[ \sum \limits_{\substack{t' \leq t\\(s_{t'},a_{t'}) \in \mathcal{D}_t}} \{ e(&s_{t'}, a_{t'}~|~\theta_t) - e(s_{t'}, a_{t'}~|~\theta_{t+1}) \} ] ~\approx~  \mathbb{E}~[ \sum \limits_{\substack{t' \leq t\\(s_{t'},a_{t'}) \in \mathcal{D}_t^+}} \left\{ e(s_{t'}, a_{t'}~|~\theta_t) - e(s_{t'}, a_{t'}~|~\theta_{t+1}) \right\} ],
\end{align}
in the limit $\lvert\mathcal{D}_t\rvert \rightarrow \infty$, since for the noisy subset $\mathcal{D}_t^-$ the improvement in prediction error after a model update $\theta_t \rightarrow \theta_{t+1}$ is zero in expectation. That is, $\mathbb{E}~[r_t(s_t, a_t~|~\mathcal{D}_t^-)] \approx 0$, or
\begin{align}
 \mathbb{E}~[ \sum \limits_{\substack{t' \leq t\\(s_{t'},a_{t'}) \in \mathcal{D}_t^-}}  e(s_{t'}, a_{t'}~|~\theta_t) ]~\approx~  \mathbb{E}~[ \sum \limits_{\substack{t' \leq t\\(s_{t'},a_{t'}) \in \mathcal{D}_t^-}} e(s_{t'}, a_{t'}~|~\theta_{t+1}) ].
\end{align}
Therefore, Eq.~\eqref{formulation1} measures the improvement in the cumulative prediction error of the learnable state transitions sampled up to time $t$ after a training step $\theta_t \rightarrow \theta_{t+1}$ on the most recent transition, in expectation.

\section{Full Derivation of the Tractable Reformulation}\label{app:derivation}

This appendix provides the full algebraic derivation of the decomposition in Eq.~\eqref{eqCT18}, which was summarised in Section~\ref{sec:subsectractableref}. Starting from Eq. \eqref{telescoping} for $C(T)$ and splitting it into two terms, we have:
\begin{align}
C(T) &=\sum \limits_{t=0}^{t=T} \gamma^t~\sum \limits_{t' \leq t} e(s_{t'}, a_{t'}~|~\theta_t) - \sum \limits_{t=0}^{t=T} \gamma^t~\sum \limits_{t' \leq t} e(s_{t'}, a_{t'}~|~\theta_{t+1}). \label{cumulativer}
\end{align}
We rewrite the second term by substituting $t = p-1$:
\begin{align}
&\sum \limits_{t=0}^{t=T} \gamma^t~\sum \limits_{t' \leq t} e(s_{t'}, a_{t'}~|~\theta_{t+1})\\
&= \sum \limits_{p=1}^{p=T+1} \gamma^{p-1}~\sum \limits_{t' \leq p-1} e(s_{t'}, a_{t'}~|~\theta_{p}) \\
&= \sum \limits_{p=1}^{p=T+1} \gamma^{p-1}~\left( -e(s_p,a_p~|~\theta_p) + \sum \limits_{t' \leq p} e(s_{t'}, a_{t'}~|~\theta_{p}) \right)\\
&=\sum \limits_{p=1}^{p=T+1} \gamma^{p-1}~\sum \limits_{t' \leq p} e(s_{t'}, a_{t'}~|~\theta_{p}) - \sum \limits_{p=1}^{p=T+1} \gamma^{p-1} e(s_p,a_p~|~\theta_p) \\
&= \left(   \sum \limits_{p=0}^{p=T+1} \gamma^{p-1}~\sum \limits_{t' \leq p} e(s_{t'}, a_{t'}~|~\theta_{p}) \right)  - \gamma^{-1} e(s_0,a_0~|~\theta_0)   - \sum \limits_{p=1}^{p=T+1} \gamma^{p-1} e(s_p,a_p~|~\theta_p) \\
&= \left(   \sum \limits_{p=0}^{p=T} \gamma^{p-1}\sum \limits_{t' \leq p} e(s_{t'}, a_{t'}~|~\theta_{p}) + \gamma^T \sum \limits_{t' \leq T+1} e(s_{t'}, a_{t'}~|~\theta_{T+1}) \right)  - \left( \gamma^{-1} e(s_0,a_0~|~\theta_0)   + \sum \limits_{p=1}^{p=T+1} \gamma^{p-1} e(s_p,a_p~|~\theta_p) \right) \\
&= \frac{1}{\gamma}\left(   \sum \limits_{p=0}^{p=T} \gamma^{p}\sum \limits_{t' \leq p} e(s_{t'}, a_{t'}~|~\theta_{p}) \right) + \gamma^T \sum \limits_{t' \leq T+1} e(s_{t'}, a_{t'}~|~\theta_{T+1})      - \sum \limits_{p=0}^{p=T+1} \gamma^{p-1} e(s_p,a_p~|~\theta_p). \\
&\text{Now, rename $p \leftarrow t$, to get:}\nonumber\\
&= \frac{1}{\gamma}\left(   \sum \limits_{t=0}^{t=T} \gamma^{t}\sum \limits_{t' \leq t} e(s_{t'}, a_{t'}~|~\theta_{t}) \right) + \gamma^T \sum \limits_{t' \leq T+1} e(s_{t'}, a_{t'}~|~\theta_{T+1})      - \sum \limits_{t=0}^{t=T+1} \gamma^{t-1} e(s_t,a_t~|~\theta_t).
\end{align}

Putting this back in Eq. \eqref{cumulativer}, we get:

\begin{align}
&C(T)\nonumber\\ &=  \left(1-\frac{1}{\gamma}\right)\left(   \sum \limits_{t=0}^{t=T} \gamma^{t}\sum \limits_{t' \leq t} e(s_{t'}, a_{t'}~|~\theta_{t}) \right) - \gamma^T \sum \limits_{t' \leq T+1} e(s_{t'}, a_{t'}~|~\theta_{T+1})    +  \sum \limits_{t=0}^{t=T+1} \gamma^{t-1} e(s_t,a_t~|~\theta_t)\\
 &=  \left(1-\frac{1}{\gamma}\right)\left(   \sum \limits_{t=0}^{t=T} \gamma^{t}\sum \limits_{t' < t} e(s_{t'}, a_{t'}~|~\theta_{t}) \right) + \left(1-\frac{1}{\gamma}\right)\left(   \sum \limits_{t=0}^{t=T} \gamma^{t} e(s_{t}, a_{t}~|~\theta_{t}) \right) - \gamma^T \sum \limits_{t' \leq T+1} e(s_{t'}, a_{t'}~|~\theta_{T+1})  +  \sum \limits_{t=0}^{t=T+1} \gamma^{t-1} e(s_t,a_t~|~\theta_t)
 \\
 &=  \left(1-\frac{1}{\gamma}\right)\left(   \sum \limits_{t=0}^{t=T} \gamma^{t}\sum \limits_{t' < t} e(s_{t'}, a_{t'}~|~\theta_{t}) \right) + \left(1-\frac{1}{\gamma}\right)\left(   \sum \limits_{t=0}^{t=T} \gamma^{t} e(s_{t}, a_{t}~|~\theta_{t}) \right)  +  \sum \limits_{t=0}^{t=T+1} \gamma^{t-1} e(s_t,a_t~|~\theta_t) - \gamma^T \sum \limits_{t' \leq T+1} e(s_{t'}, a_{t'}~|~\theta_{T+1})
  \\
 &=  \left(1-\frac{1}{\gamma}\right)\left(   \sum \limits_{t=0}^{t=T} \gamma^{t}\sum \limits_{t' < t} e(s_{t'}, a_{t'}~|~\theta_{t}) \right) + \left(1-\frac{1}{\gamma}\right)\left(   \sum \limits_{t=0}^{t=T} \gamma^{t} e(s_{t}, a_{t}~|~\theta_{t}) \right)  +  \sum \limits_{t=0}^{t=T} \gamma^{t-1} e(s_t,a_t~|~\theta_t) - \gamma^T \sum \limits_{t' \leq T} e(s_{t'}, a_{t'}~|~\theta_{T+1})
   \\
 &=  \left(1-\frac{1}{\gamma}\right)\left(   \sum \limits_{t=0}^{t=T} \gamma^{t}\sum \limits_{t' < t} e(s_{t'}, a_{t'}~|~\theta_{t}) \right) +   \sum \limits_{t=0}^{t=T} \gamma^{t} e(s_{t}, a_{t}~|~\theta_{t})    - \gamma^T \sum \limits_{t' \leq T} e(s_{t'}, a_{t'}~|~\theta_{T+1})
    \\
 &=  \left(1-\frac{1}{\gamma}\right)\left(   \sum \limits_{t=0}^{t=T} \gamma^{t}\sum \limits_{t' < t} e(s_{t'}, a_{t'}~|~\theta_{t}) \right) + \frac{1}{\gamma}  \sum \limits_{t=0}^{t=T} \gamma^{t+1} e(s_{t}, a_{t}~|~\theta_{t})    - \frac{1}{\gamma}  \sum \limits_{t' \leq T} \gamma^{T+1} e(s_{t'}, a_{t'}~|~\theta_{T+1}),
\end{align}
which is exactly Eq.~\eqref{eqCT18} of the main paper after relabelling $t' \to t$.

\section{Upper Bound and Slack Condition}\label{app:upperbound}

As stated in Section~\ref{sec:subsectractableref}, the intractable history term in Eq.~\eqref{eqCT18} is negative because the coefficient $(1-1/\gamma)$ is negative for $\gamma \in (0,1)$. So $C(T)$ in Eq.~\eqref{eqCT18} is in fact \emph{upper-bounded} by the tractable per-step term:
\begin{align*}
C(T) \;\leq\; \frac{1}{\gamma}\sum \limits_{t=0}^{t=T}\Big[\gamma^{t+1}\, e(s_t,a_t~|~\theta_t) - \gamma^{T+1}\, e(s_t,a_t~|~\theta_{T+1})\Big], \qquad \gamma \in (0,1),
\end{align*}
with slack $-(1-1/\gamma)\sum_{t=0}^{T} \gamma^{t}\sum_{t' < t} e(s_{t'},a_{t'}~|~\theta_t) \geq 0$ that vanishes as $\gamma \to 1$. This is structurally \textit{analogous to a variational upper bound} on an intractable objective: the $\gamma \to 1$ limit doubles as a \textit{tightness condition} -- the unique value at which the tractable per-step surrogate term equals $C(T)$ exactly.

\section{Telescoping Intuition}\label{app:telescoping}

The collapse of Eq.~\eqref{telescoping} to a per-step reward in Eq.~\eqref{wow1} is, at its core, a \textit{telescoping sum} in the model parameters. Setting $\gamma=1$ in Eq.~\eqref{telescoping} for $C(T)$ and re-summing the lower-triangular region $\{(t,t') : 0 \leq t' \leq t \leq T\}$ column-wise instead of row-wise (i.e., for fixed $t'$, letting $t$ range over $\{t',t'+1,\ldots,T\}$),
\begin{align*}
\sum \limits_{t=0}^{T}\sum \limits_{t' \leq t}\bigl[e(s_{t'},a_{t'}~|~\theta_t) - e(s_{t'},a_{t'}~|~\theta_{t+1})\bigr]
= \sum \limits_{t'=0}^{T}\sum \limits_{t=t'}^{T}\bigl[e(s_{t'},a_{t'}~|~\theta_t) - e(s_{t'},a_{t'}~|~\theta_{t+1})\bigr],
\end{align*}
the inner sum telescopes in $t$ -- successive $\theta_t$ and $\theta_{t+1}$ terms share the transition $(s_{t'},a_{t'})$ and cancel pairwise -- leaving $e(s_{t'},a_{t'}~|~\theta_{t'}) - e(s_{t'},a_{t'}~|~\theta_{T+1})$, and relabelling $t' \to t$ recovers Eq.~\eqref{wow1}. Intuitively: each transition's contribution to cumulative historical error improvement is simply its \emph{current-visit} error minus its \emph{end-of-training} error. This is why an $\mathcal{O}(t)$-wide history collapses to a per-step signal without any approximations.

\section{Per-step Prediction Error Rewards as Special Cases}\label{app:specialcases}

Schmidhuber's formulations of prediction error-based curiosity, namely, Curiosity V1 \cite{CV1} and Curiosity V2 \cite{CV2}, naturally fit into the Curiosity-Critic framework.

For instance, when the environment is assumed to be deterministic, the expected asymptotic error baseline for a transition $(s_t,a_t)\rightarrow s_{t+1}$ is $\mathbb{E}_{\mathcal{P}}\left[ e(s_t, a_t~|~\theta_{\infty})\right]=0$, as a converged world model can ideally predict the state transition perfectly. Under this condition, our proposed reward in Eq. \eqref{effrewardv3opt} reduces to the Curiosity V1 \cite{CV1} reward with $r^{\text{eff}}_t(s_t,a_t) = e(s_t, a_t ~|~\theta_t)$.

Now, consider the case where the expected asymptotic error baseline for a transition $\mathbb{E}_{\mathcal{P}}[ e(s_t, a_t~|~\theta_{\infty})]$ is approximated by the error in the immediate future after one training update, i.e., $\mathbb{E}_{\mathcal{P}}[ e(s_t, a_t~|~\theta_{\infty})] \approx e(s_t, a_t~|~\theta_{t+1})$. Under this condition, our proposed reward in Eq. \eqref{effrewardv3opt} reduces to the Curiosity V2 \cite{CV2} reward with $r^{\text{eff}}_t(s_t,a_t) = e(s_t, a_t ~|~\theta_t) - e(s_t, a_t~|~\theta_{t+1})$.

Both reductions hold for any valid notion of prediction error $e(\cdot)$, so they apply beyond Schmidhuber's original $L_2$-on-observations formulation. In particular, any Curiosity V1 variant obtained by choosing a different error metric, such as ICM's squared $L_2$ in a learned inverse-dynamics feature space \cite{pathak1_icm}, is also a special case of our framework under the zero-error-baseline approximation.

\section{Jensen's Inequality Bound on the Asymptotic Error Baseline}\label{app:jensen}

Using Jensen's inequality, one can also note that the L2 error baseline in Eq. \eqref{opterrorform} is upper bounded by the square-root of the summed variance of the state transition:
\begin{align}
\mathbb{E}_{\mathcal{P}}\left[ \| \mathbb{E}_{\mathcal{P}}\left[ s_{t+1} \right] - s_{t+1} \| \right] \leq \sqrt{\mathbb{E}_{\mathcal{P}}\left[ \| \mathbb{E}_{\mathcal{P}}\left[ s_{t+1} \right] - s_{t+1} \|^2 \right]} = \sqrt{\sum_i \text{Var}_\mathcal{P}(s^{(i)}_{t+1})},
\end{align}
where $i$ indexes the dimensions of the state $s_{t+1}$.

\section{Curiosity-Critic Algorithm}\label{app:algorithm}

\begin{algorithm}[H]
\caption{Curiosity-Critic (online, batch-size-one instantiation)}
\label{alg:cc}
\begin{algorithmic}[1]
\REQUIRE World model $\theta_t$, neural critic $\phi_t$, policy $\pi_t$, current state $s_t$
\STATE Sample action $a_t \sim \pi_t(\cdot \mid s_t)$; execute and observe $s_{t+1}$
\STATE $e_{\text{before}} \leftarrow e(s_t, a_t \mid \theta_t)$ \hfill \textit{// world model error before update}
\STATE $\theta_{t+1} \leftarrow \mathrm{GradStep}(\theta_t;\ s_t, a_t, s_{t+1})$ \hfill \textit{// update world model}
\STATE $e_{\text{after}} \leftarrow e(s_t, a_t \mid \theta_{t+1})$ \hfill \textit{// post-update error}
\STATE $\phi_{t+1} \leftarrow \mathrm{GradStep}(\phi_t;\ s_t, a_t,\ e_{\text{after}})$ \hfill \textit{// update critic to predict }$e_{\text{after}}$
\STATE $r_t \leftarrow e_{\text{before}} - \phi_{t+1}(s_t, a_t)$ \hfill \textit{// curiosity reward (Eq.~\ref{effrewardv3opt})}
\STATE $\pi_{t+1} \leftarrow \mathrm{PolicyUpdate}(\pi_t;\ s_t, a_t, r_t)$ \hfill \textit{// update policy with }$r_t$
\STATE \textbf{return~} $\theta_{t+1},\ \phi_{t+1},\ \pi_{t+1},\ s_{t+1}$
\end{algorithmic}
\end{algorithm}

\textbf{Remark (batch size and update cadence).}
Algorithm~\ref{alg:cc} is the per-step, batch-size-one instantiation used in our
grid world, where $\theta$ and $\phi$ are updated once per environment step. The
framework requires neither: the $\mathrm{GradStep}$ on lines~3 and~5 may be a
minibatch or replay-batch update, and the reward on line~6 may be computed from
frozen $\theta,\phi$ snapshots over a batch of transitions \emph{before} the update,
as is standard for on-policy deep RL. In a PPO instantiation, for example, the
intrinsic rewards for a full rollout are computed with frozen $\theta,\phi$; $\theta$
is then updated on minibatches, the post-update error $e_{\text{after}}$ is evaluated
on those same minibatches, and $\phi$ regresses it, so the next rollout uses the
updated models.

\section{Environment and Evaluation Details}\label{app:envdetails}

\textbf{Environment.}
The environment is a $30 \times 30$ grid world divided into two equal regions.

The \emph{deterministic region} occupies the entire left half (all rows, columns 0--14), comprising $450$ cells, each assigned a fixed 200-dimensional binary observation pattern (``TV pixels''), constructed as cyclic shifts of a shared random base vector such that neighbouring cells have strongly correlated observations.

The \emph{stochastic region} occupies the right half (columns 15--29); every visit independently re-samples an i.i.d.\ Bernoulli$(0.5)$ 200-dimensional binary vector, making prediction error irreducible regardless of training.
The oracle irreducible noise floor for the stochastic region is $\sqrt{200} \times 0.5 \approx 7.07$ (and trivially $0$ for the deterministic region).
The agent begins at cell $(15, 15)$ --- the centre row, on the stochastic side of the boundary.

\textbf{Actions and transitions.}
At each step the agent selects one of four directional actions: Up, Down, Left, Right.
Transitions are fully deterministic; boundary actions are excluded from the valid set at edge cells.

\textbf{World model.}
Each cell $(r, c)$ is represented by a factored 60-dimensional input: the concatenation of one-hot$(r)$ and one-hot$(c)$ of length 30 each.
The world model predicts the TV-pixel observation at the current cell (not the next state, since all transitions are trivially deterministic). The Curiosity-Critic framework places no constraint on what the world model predicts -- the theory requires only a consistent notion of prediction error, which here is the L2 error in predicting the current cell's observation.
This design isolates each curiosity signal's ability to direct the agent toward learnable regions and away from irreducibly stochastic ones - the ``noisy TV'' problem of Burda et al.\ \cite{pathak1} - in a controlled grid world setting.

\textbf{Evaluation.}
The primary metric is the mean L2 prediction error of the world model over all 450 deterministic cells, queried directly every 100 environment steps.
A lower value reflects both faster convergence and more accurate world-model learning in the learnable region.
All results are averaged over five independent seeds (seeds 1--5).
The Visitation Count reward follows Strehl \& Littman \cite{Visitation} and Bellemare et al.\ \cite{bellemare2016}: $r(s) = 1/\sqrt{N(s)}$, with $N(s)$ initialised to 1 per state.

\section{Training Details}\label{app:training}

\textbf{World model.}
The world model $\theta$ is a two-layer MLP:
\[
\text{60 (input)} \;\rightarrow\; 1024 \;\xrightarrow{\text{ReLU}}\; \text{200 (output)},
\]
trained with MSE loss using Adam \cite{adam} ($\text{lr}=0.001$, $\beta_1=0.9$, $\beta_2=0.999$).
One gradient step is taken per environment step (the minimal online setting; our framework also admits minibatch updates -- see Appendix~\ref{app:algorithm}).
Before the main loop, the model is warmed up for 100 steps under a random policy, a cold start common to all methods.

\textbf{Curiosity-Critic.}
The \emph{Neural Critic Model} $\phi$ is a small MLP,
\[
\text{60 (input)} \;\rightarrow\; 128 \;\xrightarrow{\text{ReLU}}\; \text{1 (output)},
\]
trained online with MSE loss to predict post-update world model error $e(\cdot | \theta_{t+1})$ at each visited state (one Adam step per environment step, $\text{lr}=0.001$).
Because the critic shares the same factored one-hot encoding as the world model, it generalises its baseline estimate across cells sharing a row or column, building a reliable noise-floor estimate with fewer per-state visits.
As an ablation, we also evaluate a \emph{Tabular Critic}: a $30\!\times\!30$ lookup table of per-state EMA-mean values (decay $0.9$) of the post-update world model error as a parameter-free alternative to the MLP critic.

\textbf{Random Network Distillation.}
For RND \cite{rnd}, a fixed random target network and a trainable predictor network share the architecture
\[
\text{input} \;\rightarrow\; 128 \;\xrightarrow{\text{ReLU}}\; 128,
\]
and the intrinsic reward is the predictor's pre-update MSE to the target output on the current input. The predictor is then trained for one Adam step ($\text{lr}=0.001$) on that same input. We report two input choices because our grid world has deterministic state transitions but stochastic observations: \emph{RND (State)} uses the 60-dimensional factored row/column encoding, while \emph{RND (Observation)} uses the 200-dimensional TV-pixel observation.

\textbf{Policy.}
A state-only V-table of size $30 \times 30$ tracks an exponential moving average of received intrinsic rewards per state and the agent acts greedily in V over valid neighbours:
\[
V(s) \;\leftarrow\; V(s) + \alpha\,\bigl[r_{t} - V(s)\bigr],
\]
with step size $\alpha = 0.05$. The V-table is uniformly initialised to $3.0$ across all methods and seeds. This eliminates initial optimism asymmetries, ensuring that observed differences in learning dynamics are attributable solely to the intrinsic reward signal.

At each step, with probability $\varepsilon = 0.3$ a random valid action is sampled uniformly; otherwise the agent selects the action that maximises $V$ among its neighbours:
\[
a_t = \arg\max_{a \in \mathcal{A}(s_t)} V\!\bigl(\mathrm{step}(s_t, a)\bigr),
\]
where $\mathcal{A}(s_t)$ denotes the set of valid actions at state $s_t$.

\textbf{Reward clipping.}
Given our choice of the L2 error metric for demonstrating curiosity-based exploration, we clip the intrinsic reward to be non-negative, mapping any
negative value to $0$ prior to the policy update. That is, with $e(\cdot)$ taken as the L2
prediction error, the effective reward
$e(s_t, a_t~|~\theta_t) - \mathbb{E}_{\mathcal{P}}[e(s_t, a_t~|~\theta_\infty)]$
is non-negative in expectation, since the current error cannot fall below the
asymptotic noise floor. Negative values therefore arise only as transient artifacts
of a noisy single-sample or early-training estimate, and clipping keeps
these from injecting spurious negative rewards into the policy. This matters only
for rewards that subtract a baseline and can momentarily go negative, namely
Curiosity V2 \cite{CV2} and all Curiosity-Critic variants (Neural, Tabular, and
Oracle). For rewards that are non-negative by construction, such as Curiosity V1
\cite{CV1}, Visitation Count \cite{Visitation}, and RND
\cite{rnd}, it has no effect.

\textbf{Reward normalization.}
Before each V-table update, intrinsic rewards are divided by a running exponential moving average (EMA) estimate of their standard deviation (without mean subtraction, to preserve sign and positivity).
This decouples reward magnitude from the learning rate, ensuring a fair comparison across methods with very different reward scales.

\textbf{Oracle reference.}
The Oracle Critic uses the known analytical irreducible error as the baseline: $\sqrt{200}\times 0.5$ for stochastic cells and $0$ for deterministic cells.
This requires privileged knowledge of cell type and noise level unavailable in practice.
It is included solely as a reference ceiling on what any critic-based approach can achieve with a perfect error baseline estimate.

\textbf{Compute resources.}
All experiments were run on a single 15-inch Apple MacBook (M4, 2025) with 16\,GB unified memory. No GPU or cluster resources were used.

\section{Detailed Per-Method Analysis}\label{app:permethod}

This appendix expands the per-method results summarised in Section~\ref{sec:experiments}.

\textbf{Curiosity V1 \cite{CV1}: complete failure in the presence of noise.}
Curiosity V1 \cite{CV1} assigns intrinsic reward equal to the raw prediction error, which is irreducibly high for stochastic cells regardless of how much training has occurred.
The agent rapidly converges to the noisy half of the grid: the deterministic-region visit fraction falls from $11.2\%$ in early training to essentially $0\%$ by step 15{,}000 and remains there for the rest of the run.
The mean error never drops below $6.8$ across any individual seed and finishes at $7.114 \pm 0.147$ -- indistinguishable from an untrained model evaluated on deterministic cells.
This is the so-called noisy-TV failure \cite{pathak1,pathak1_icm}.

\textbf{Visitation Count \cite{Visitation}.}
The count-based bonus $1/\sqrt{N(s)}$ generates broad early coverage (deterministic-region fraction of $54.6\%$ in early training and $58.7\%$ in late training), which might naively suggest effective exploration.
However, the bonus decays equally for every visited cell regardless of whether it is learnable, so the agent continues to allocate substantial time to stochastic cells throughout training.
The world model can only improve on deterministic visits; the equal pressure toward stochastic cells dilutes learning, and the error plateaus between $5.5$ and $5.9$ from step 14{,}000 onward.
The final error of $5.588 \pm 0.794$ is among the highest, and the seed variance ($\pm 0.794$) is the highest among all methods.

\textbf{Random exploration.}
As a baseline with no intrinsic reward, the agent acts uniformly at random, reaching $2.348 \pm 0.377$ at step $35{,}000$. 

Random is a strong baseline method here: with no interior obstacles and the learnable region adjacent to the start cell, an undirected walk reaches learnable cells with high probability. This advantage is geometry-dependent. In obstacle-rich mazes that require long-horizon directed traversal, an undirected walk only rarely reaches the learnable region, so Random would be a far weaker baseline.

Under irreducible stochasticity, Curiosity V2's single-sample baseline gives spurious reward on noisy cells and can underperform diffusion -- not that Curiosity V2 is dominated by Random in general.

\textbf{Curiosity V2 \cite{CV2}.}
Its reward measures the one-step reduction in prediction error after a gradient update.
In practice this signal is highly variable: a single update can produce a large error reduction on any cell, regardless of whether that cell lies in the deterministic or stochastic region.
As a result the agent's exploration is scattered, with a deterministic-region fraction of only $14.6\%$ in early training and $32.2\%$ in late training.
The mean error does not drop below $3.0$ until step $34{,}400$ -- a threshold our Neural Critic reaches more than $21{,}000$ steps earlier -- and the run finishes at $2.939 \pm 0.398$, worse than random exploration on average despite substantially improving over Curiosity V1.

\textbf{RND (Observation) \cite{rnd}.}
When RND receives the observed TV-pixel vector, each visit to a stochastic cell produces one newly sampled Bernoulli observation vector. The predictor therefore sees persistent input novelty in precisely the region where the world model has irreducible error, producing the same qualitative noisy-TV failure as Curiosity V1. The deterministic-region visit fraction falls from $22.3\%$ in early training to $1.3\%$ in the final $5{,}000$ steps, and the final deterministic-cell error remains high at $6.842 \pm 0.212$.

\textbf{RND (State) \cite{rnd}.}
When RND receives the deterministic 60-dimensional row/column state, the target is fixed per grid cell and the predictor error decays with repeated visits, making the method behave like a learned soft state-count bonus. This avoids the observation-level noisy-TV trap and is the strongest non-critic baseline, reaching $2.220 \pm 0.109$ with a late-training deterministic-region fraction of $52.8\%$. However, because state novelty is agnostic to whether a state's observations are learnable, RND (State) continues to spend substantial time in the stochastic half. Its mean error drops below $3.0$ at step $21{,}700$ and below $2.5$ at step $28{,}800$, but it never reaches $2.0$ within $35{,}000$ steps.

\textbf{Ours (Tabular Critic).}
As a sanity check, we evaluate a minimal critic: a $30\!\times\!30$ EMA-mean lookup table that tracks the per-state post-update error with no generalization across states.
Despite its simplicity, this variant still reaches $1.912 \pm 0.070$ for the final error.
In comparison to the Neural Critic, its slower convergence ($<3.0$ error not reached until step $25{,}300$, versus $13{,}400$ for the Neural Critic) and modest late-training deterministic-region fraction ($32.0\%$) confirm that cross-state generalization, provided by the neural critic, is key.

\textbf{Ours (Neural Critic Model).}
Replacing the per-state EMA table with a small MLP critic yields a substantial improvement in convergence speed.
The neural critic shares the same factored one-hot encoding as the world model, so a visit to one cell immediately informs the critic's estimate for other cells in the same row or column.
This generalization allows the critic to calibrate its noise-floor estimate with far fewer per-state observations, redirecting exploration much earlier. The error curve drops below $3.0$ at step $\mathbf{13{,}400}$, nearly $12{,}000$ steps ahead of the Tabular Critic, $8{,}300$ steps ahead of RND (State), and more than $21{,}000$ steps ahead of Curiosity V2 \cite{CV2}.
The error continues to fall steeply: below $2.5$ at step $16{,}000$, below $2.0$ at step $26{,}700$.
The final mean error of $\mathbf{1.858 \pm 0.080}$ is the best among all non-oracle methods. The late-training deterministic-region fraction of $70.9\%$ further confirms the neural critic's stronger directional effect on exploration: the agent spends nearly three-quarters of the final $5{,}000$ steps in the learnable half of the grid, compared to $52.8\%$ for RND (State) and $42.6\%$ for random exploration.

\textbf{Oracle upper bound.}
The ideal-critic variant, which substitutes the true analytical irreducible error floor, achieves the strongest late-training focus ($95.3\%$ deterministic-region fraction) and the lowest final error ($1.736 \pm 0.063$).
The $\sim\!7\%$ gap between the neural curiosity-critic method and the oracle in the final world model error represents the residual gap in the neural critic's estimate of the noise floor.
The neural critic serves as a practical approximation of the irreducible error floor, requiring no privileged knowledge of cell type or noise level.

\section{Additional Figures}\label{app:figures}

\begin{figure}[h]
    \centering
    \includegraphics[width=\textwidth]{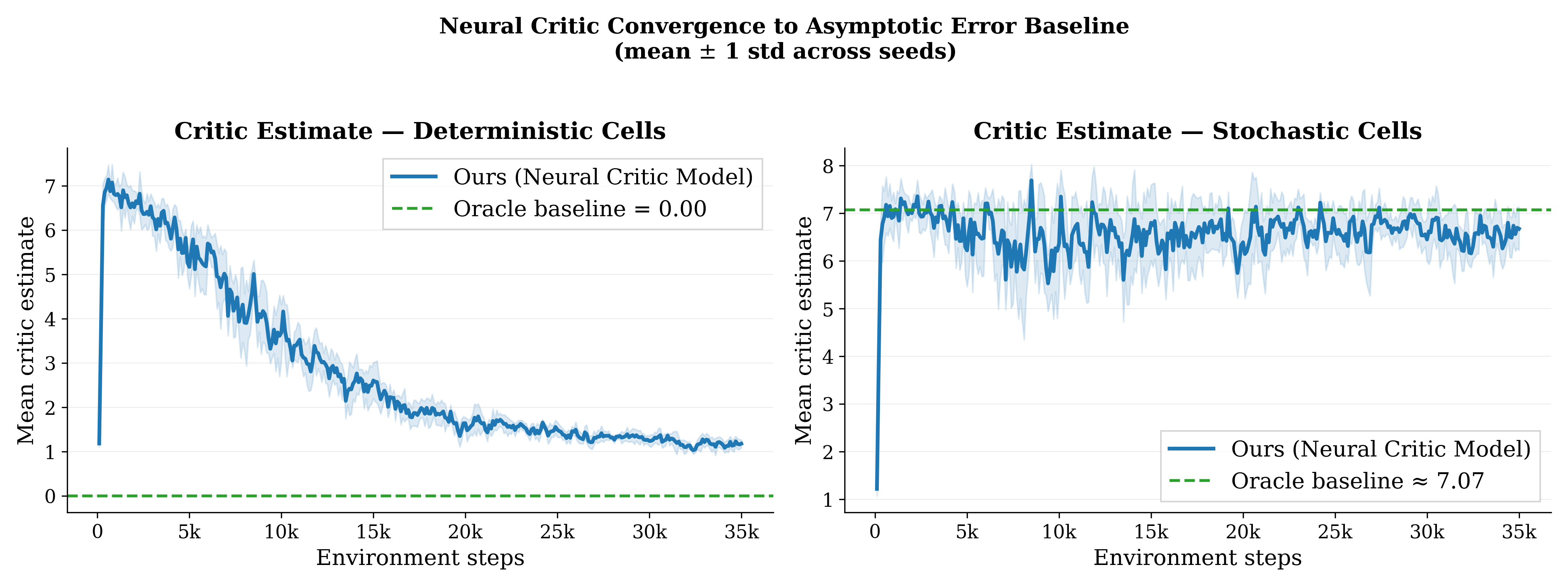}
    \caption{Mean neural critic estimate over all deterministic cells (left) and all stochastic cells (right) versus environment steps, averaged over five seeds with $\pm 1$ standard deviation shading.
    The dashed horizontal line marks the oracle noise floor ($0$ for deterministic, $\sqrt{200}\times 0.5 \approx 7.07$ for stochastic).
    The critic estimate for stochastic cells first reaches within $5\%$ of the oracle by step $400$; the critic estimate for deterministic cells rises in the first $\sim\!500$ steps (tracking the large world model error early in training) and then declines as the world model learns the learnable region.}
    \label{fig:critic_convergence}
\end{figure}

\begin{figure}[h]
    \centering
    \includegraphics[width=\textwidth]{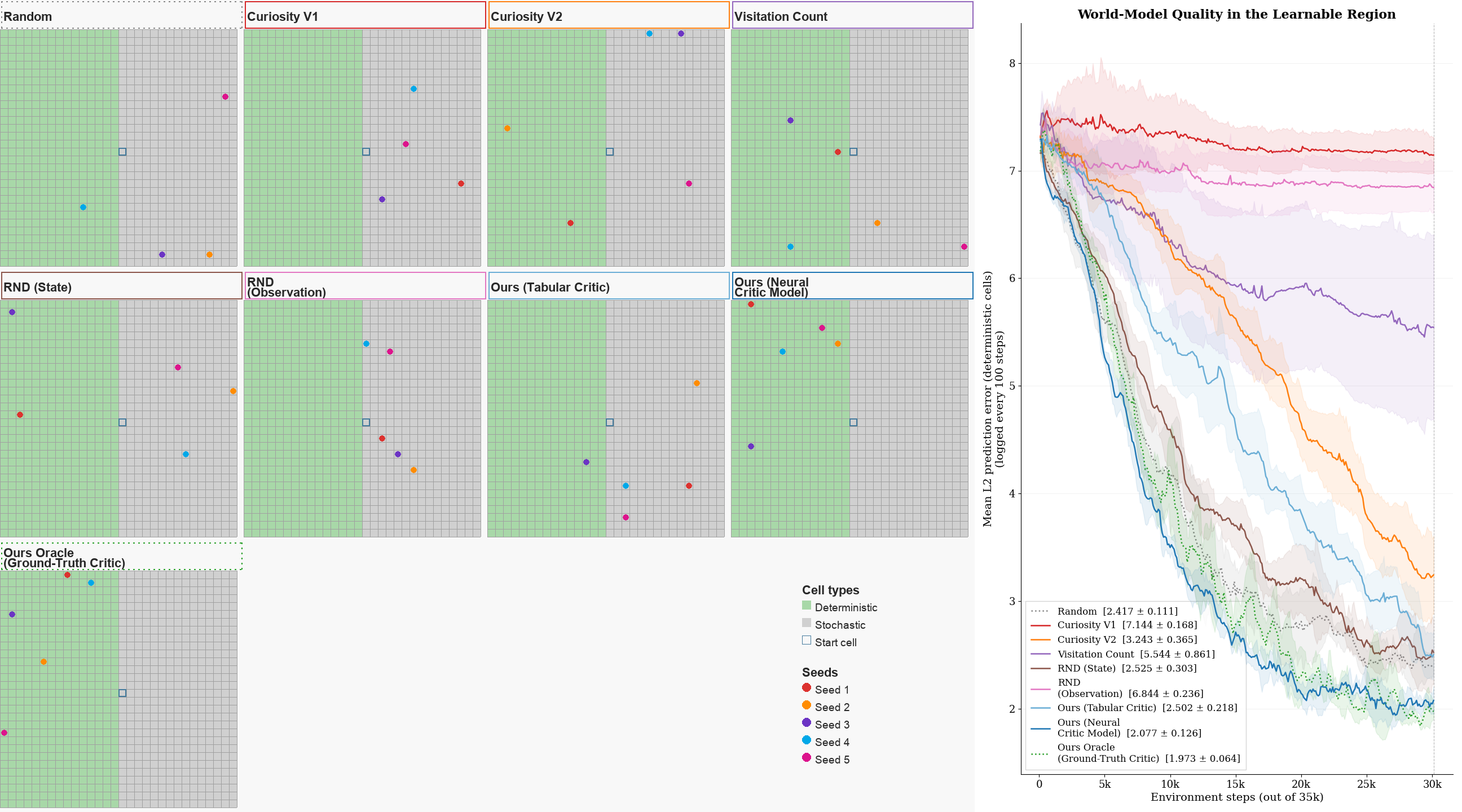}
    \caption{Snapshot of agent trajectories at environment step 30{,}000.
    Each panel shows one method; coloured blobs represent individual seeds (five seeds per method, simultaneously animated).
    Green cells are deterministic (learnable); grey cells are stochastic (irreducibly noisy).
    The rightmost column shows the live mean L2 prediction error trace up to this step.
    Curiosity V1 \cite{CV1} (prediction error as intrinsic reward) and Curiosity V2 \cite{CV2} (one-step learning progress as intrinsic reward) are shorthands used throughout; RND (State) and RND (Observation) distinguish whether the RND input is the deterministic grid state or stochastic TV-pixel observation. A full real-time animation is available at \href{https://youtu.be/hHSvQGaO5yY}{\texttt{youtu.be/hHSvQGaO5yY}}.}
    \label{fig:snapshot}
\end{figure}

\begin{figure}[h]
    \centering
    \includegraphics[width=0.92\textwidth]{heatmap_end.png}
    \caption{Seed-averaged visitation heatmaps at the end of training (final 5{,}000 of 35{,}000 steps), shown for the displayed methods.
    Each cell's brightness encodes the fraction of time spent there, normalised per method.
    The cyan dashed line marks the boundary between the deterministic region (left) and the stochastic region (right).
    Ours (Neural Critic Model) concentrates late-training visits in the deterministic half, close to the oracle; Curiosity V1 \cite{CV1} and RND (Observation) concentrate in the stochastic half; Random and RND (State) produce broader coverage, with RND (State) less targeted than the critic-based methods.}
    \label{fig:heatmap_end}
\end{figure}

\begin{figure}[h]
    \centering
    \includegraphics[width=\textwidth]{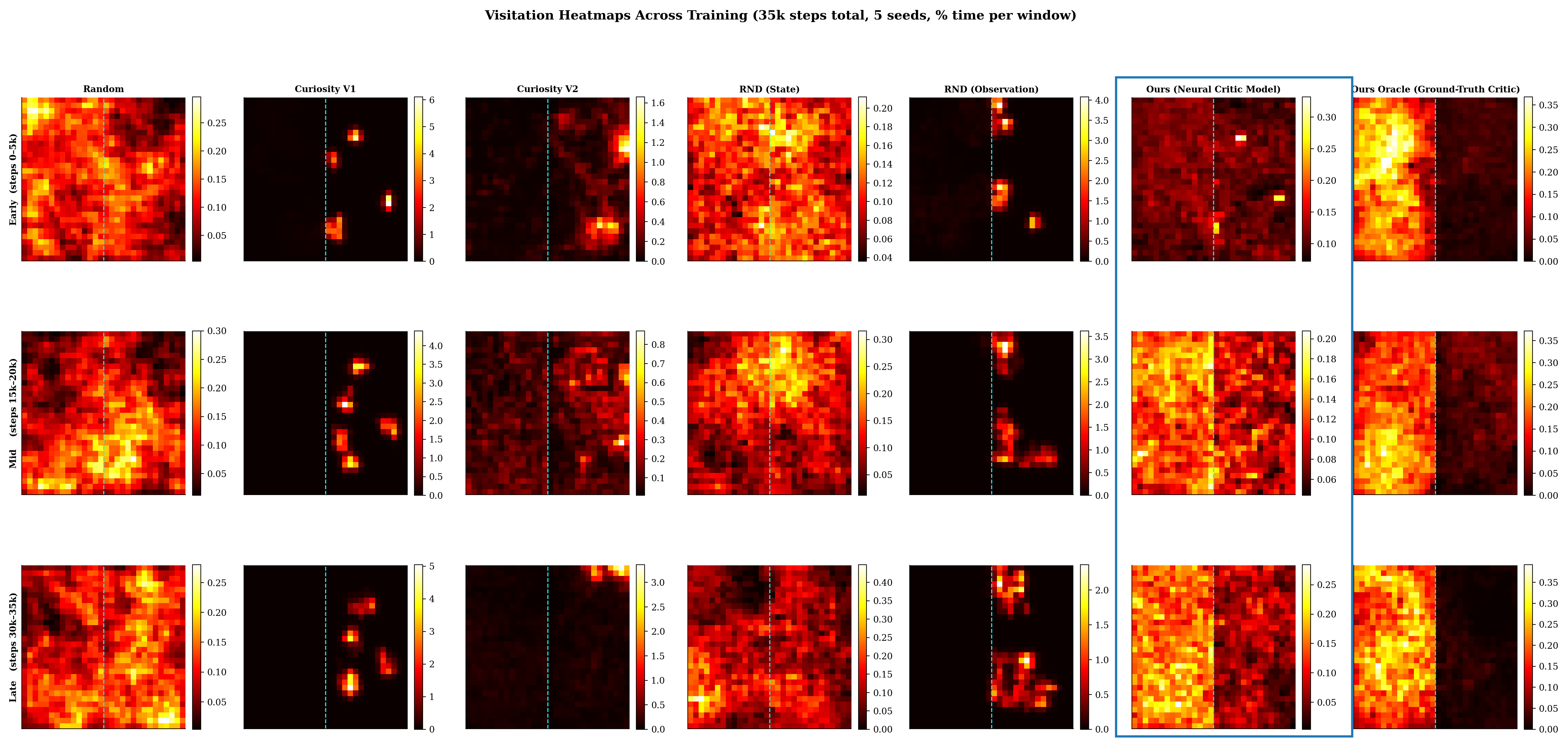}
    \caption{Seed-averaged visitation heatmaps at three training windows: Early (steps 0--5k), Mid (steps 15k--20k), and Late (steps 30k--35k), shown for the displayed methods.
    Rows correspond to training windows; columns to methods.
    The highlighted column (blue outline) shows Ours (Neural Critic Model), which exhibits a clear shift from broad early exploration to targeted late-training concentration in the deterministic region.
    The oracle critic (rightmost) maintains high deterministic-region focus throughout.
    RND (Observation) and Curiosity V1 \cite{CV1} remain fixated on the stochastic region, while RND (State) retains broader state-novelty-driven coverage without the critic's targeted late-training focus.
    Curiosity V1 and V2 are convenience shorthands for \cite{CV1} and \cite{CV2} respectively.}
    \label{fig:heatmap_windows}
\end{figure}

%%%%%%%%%%%%%%%%%%%%%%%%%%%%%%%%%%%%%%%%%%%%%%%%%%%%%%%%%%%%%%%%%%%%%%%%%%%%%%%
%%%%%%%%%%%%%%%%%%%%%%%%%%%%%%%%%%%%%%%%%%%%%%%%%%%%%%%%%%%%%%%%%%%%%%%%%%%%%%%

\end{document}